\documentclass[a4paper,twoside]{article}

\usepackage{epsfig}
\usepackage{subcaption}
\usepackage{calc}
\usepackage{amssymb}
\usepackage{amstext}
\usepackage{amsmath}
\usepackage{amsthm}
\usepackage{multicol}
\usepackage{pslatex}
\usepackage{apalike}
\usepackage[bottom]{footmisc}
\usepackage{SCITEPRESS}     % Please add other packages that you may need BEFORE the SCITEPRESS.sty package.

\usepackage[dvipsnames]{xcolor}

\begin{document}

\title{Insertion of real agents behaviors in CARLA autonomous driving simulator}

\author{\authorname{Sergio Martín Serrano\sup{1}\orcidAuthor{0000-0002-8029-1973}, David Fernández Llorca \sup{1, 2}\orcidAuthor{ 0000-0003-2433-7110}, Iván García Daza\sup{1}\orcidAuthor{0000-0001-8940-6434} \\ and Miguel Ángel Sotelo\sup{1}\orcidAuthor{ 0000-0001-8809-2103 }}
\affiliation{\sup{1}Computer Engineering Department, University of Alcalá, Alcalá de Henares, Spain}
\affiliation{\sup{2}European Commission, Joint Research Centre (JRC), Seville}
\email{\{sergio.martin, david.fernandezl, ivan.garciad, miguel.sotelo\}@uah.es}
}

\keywords{Automated driving, autonomous vehicles, predictive perception, behavioural modelling, simulators, virtual reality, presence.}

\abstract{The role of simulation in autonomous driving is becoming increasingly important due to the need for rapid prototyping and extensive testing. The use of physics-based simulation involves multiple benefits and advantages at a reasonable cost while eliminating risks to prototypes, drivers and vulnerable road users. However, there are two main limitations. First, the well-known \emph{reality gap} which refers to the discrepancy between reality and simulation that prevents simulated autonomous driving experience from enabling effective real-world performance. Second, the lack of empirical knowledge about the  \emph{behavior of real agents}, including backup drivers or passengers and other road users such as vehicles, pedestrians or cyclists. Agent simulation is usually pre-programmed deterministically, randomized probabilistically or generated based on real data, but it does not represent behaviors from real agents interacting with the specific simulated scenario. In this paper we present a preliminary framework to enable real-time interaction between real agents and the simulated environment (including autonomous vehicles) and generate synthetic sequences from simulated sensor data from multiple views that can be used for training predictive systems that rely on behavioral models. Our approach integrates immersive virtual reality and human motion capture systems with the CARLA simulator for autonomous driving. We describe the proposed hardware and software architecture, and discuss about the so-called \emph{behavioural gap} or \emph{presence}. We present preliminary, but promising, results that support the potential of this methodology and discuss about future steps. }

\onecolumn \maketitle \normalsize \setcounter{footnote}{0} \vfill

\section{\uppercase{Introduction}}
\label{sec:introduction}

\begin{figure*}[h]
  \centering
  \includegraphics[width=0.7\linewidth]{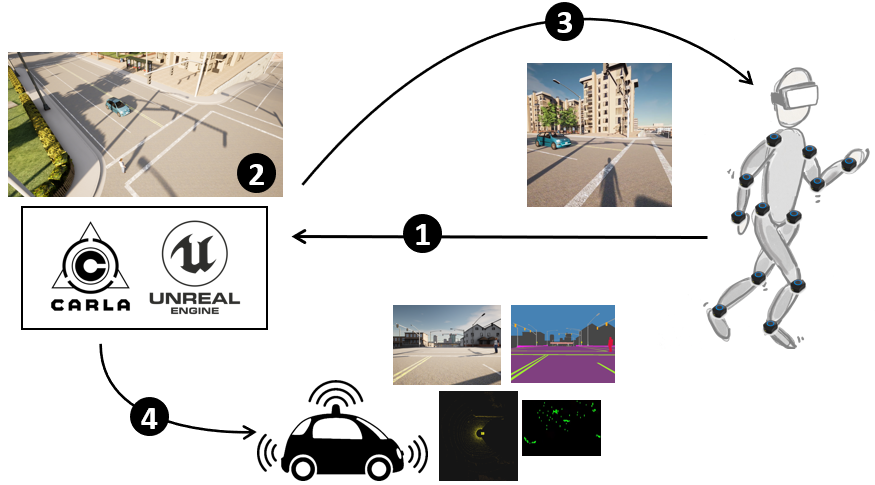}
  \caption{Overview of the presented approach. (1) CARLA-Unreal Engine is provided with the head (VR headset) and body (motion capture system) pose. (2) The scenario is generated, including the autonomous vehicles and the digitized pedestrian. (3) The environment is provided to the pedestrian (through VR headset). (4) Autonomous vehicle sensors perceive the environment, including the pedestrian. }
  \label{figure_1}
\end{figure*}

The recent explosion in the use of simulators for rapid prototyping and testing of autonomous systems has occurred primarily in the field of autonomous vehicles. This is mostly due to the compelling need for extensive validation, as real test-driving alone cannot provide sufficient evidence for demonstrating their safety \cite{Kalra2016}. The relative ease of generating massive data for training and testing, especially edge cases, having full control over all the variables under study (e.g., street layout, lighting conditions, traffic scenarios), makes its use very attractive. In addition, the generated data is annotated by design including semantic information.  

But the use of virtual-world data also brings with it some challenges. The realism of the simulated sensor data (i.e., radar, LiDAR, cameras) and the physical models are critical aspects to minimize the reality gap. Advances in this area are very remarkable, and highly realistic synthetic data has been shown to be useful for improving different perception tasks such as pedestrian detection \cite{Vazquez2014}, semantic segmentation \cite{Ros2016} or vehicle speed detection \cite{Hernandez2021}. However, despite efforts to generate realistic synthetic behaviors of road agents (e.g., to address trajectory forecasting \cite{Weng2020_AIODrive}), simulation lacks empirical knowledge about the behavior of other road users (e.g., vehicles, pedestrians, cyclists), including behavior and motion prediction, communication, and human-vehicle interaction \cite{Eady2019}. 

In this paper, we present a new approach to incorporate real agents behaviors and interactions in CARLA autonomous driving simulator \cite{carla2017} by using immersive virtual reality and human motion capture systems. The idea is to integrate a subject in the simulated scenarios using CARLA and Unreal Engine 4 (UE4), with real time feedback of the pose of his head and body, and including positional sound, trying to generate an experience realistic enough for the subject to feel physically present in the virtual world and accept it at a subconscious level (i.e., maximize virtual reality presence). At the same time, the captured pose and motion of the subject is integrated into the virtual scenario by means of an avatar, so that the simulated sensors of the autonomous vehicles (i.e., radar, LiDAR, cameras) can be aware of their presence. This is very interesting because it allows, on the one hand, to obtain synthetic sequences from multiple points of view based on the behavior of real subjects, which can be used to train and test predictive perception models. And on the other hand, they also allow to address different types of interaction studies between autonomous vehicles (i.e., software in the loop) and real subjects, including external human-machine interfaces (eHMI), in fully controlled conditions and with total safety. To our knowledge, this is the first attempt to integrate the behaviors of real road agents into an autonomous vehicle simulator for these purposes.  

\begin{figure*}[ht]
  \centering
  \includegraphics[width=0.99\linewidth]{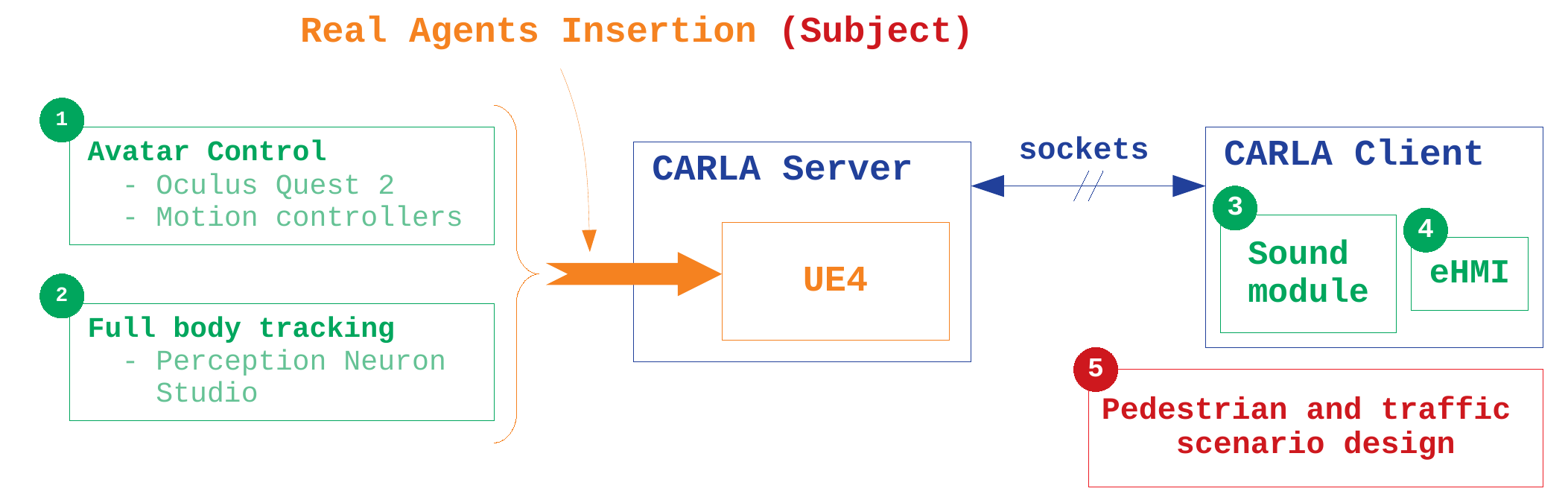}
  \caption{System Block Diagram.}
  \label{fig_Block_Diagram}
\end{figure*}

\section{\uppercase{RELATED WORK}}

In the following, we summarize the most relevant works that have proposed the use of virtual reality and simulators for vulnerable road users (mainly pedestrians and cyclists) in the context of automated driving. Since the visual realism of simulators has improved greatly in recent years, we focus on relatively recent work, as the conclusions obtained with less realistic simulators are not entirely applicable to the current context. 

Most of the previous works proposed the use of virtual reality and simulators to study pedestrian behavior in scenarios where they have to interact with an autonomous vehicle (high levels of automation \cite{Llorca2021}) including the perception of distances and the intention to cross \cite{Doric2016}, \cite{Deb2017}, \cite{IryoAsano2018}, \cite{Bhagavathula2018}, \cite{Farooq2018}, \cite{NunezVelasco2019}.

Additionally, these approaches have recently been used to study the communication between autonomous vehicles and pedestrians, including acceptance and behavioral analysis of various eHMI systems \cite{Chang2017}, \cite{Hollander2019}, \cite{Shuchisnigdha2020}, \cite{Gruenefeld2019}, \cite{Locken2019}, \cite{Nguyen2019}. Virtual environments allow experimentation with multiple types of eHMI systems, and with multiple types of vehicles in a relatively straightforward fashion. 

We can find similar studies for the case of cyclists, based on simulation platforms and immersive virtual reality \cite{Ullmann2020}, usually to analyze the perceptions and behavior of cyclists to various environmental variables \cite{Xu2017}, \cite{NunezVelasco2021}, \cite{Nazemi2021}. 

The aforementioned studies have not considered the use of motion capture systems to integrate visual body feedback into the simulation. This is an important factor for two main reasons. First, a fully-articulated visual representation of the users in the immersive virtual environment can enhance their subjective sense of feeling present in the virtual world \cite{Bruder2009}. Second, and more important for our purpose, the articulated representation of the pedestrian pose (body, head, and even hands) within the simulator allows, on the one hand, to integrate the user within the simulated scenario and to be detected by the sensors of the autonomous vehicles, allowing the generation of data sequences that can be very useful for training and testing predictive models. On the other hand, it allows real-time interaction between the autonomous vehicle and the pedestrian (i.e., software in the loop), which can be very useful to test edge cases.

In some studies, user-autonomous vehicle interaction is not possible, as they rely on the projection of prerecorded 360º videos (i.e., 360º video-based virtual reality \cite{NunezVelasco2019}, \cite{NunezVelasco2021}). While this approach certainly enhances visual realism, it does not enable interaction. But even with the use of simulators, we have found only one paper \cite{Gruenefeld2019} that attempts some kind of interaction between the pedestrian and the autonomous vehicle via hand gestures (i.e. stop gesture). However, the gesture detection process of the autonomous vehicle in the simulator does not seem to be related to the integration of the user in the virtual scenario. 

Finally, regarding open source autonomous driving simulators we find several options such as CARLA \cite{carla2017}, AirSim \cite{airsim} or DeepDrive \cite{deepdrive}, with CARLA and AirSim being probably the ones that are receiving more attention and have experienced the largest growth in recent years. Although AirSim works natively with virtual reality (originally intended as flight simulator), CARLA does not support virtual reality natively. The introduction of immersive virtual reality with CARLA has only recently been proposed for the driver use case \cite{Silvera2022}. To the best of our knowledge, this is the first work that provides a solution to integrate subjects into the CARLA simulator using immersive virtual reality for the vulnerable road users case, including motion capture, visual body feedback and integration of the articulated human pose into the simulator. 

\section{\uppercase{VR Immersion Features}}
This section describes the features of the immersive virtual reality system in the CARLA simulator for autonomous driving. Full pedestrian immersion is achieved by making use of the functionality provided by UE4 and external hardware, such as VR glasses and a set of motion sensors, for behaviour and interaction research.

The CARLA open-source simulator is implemented over UE4, which provides high rendering quality, realistic physics and an ecosystem of interoperable plugins. 
%CARLA simulates a dynamic world, \ig{It is not clear the sentence} providing an interface between the world created by UE4 and several agents that interact with the world. 
CARLA simulates dynamic traffic scenarios and provides an interface between the virtual world created with UE4 and the road agents operating within the scenario. 
CARLA is designed as a server-client system to make this possible, where the server runs the simulation and renders the scene. 
%\ig{Try to clarify the following sentence}. The client API is coded in Python and allows interaction between client-generated agents and the server via sockets.  
Communication between the client and the server is done via sockets.

%\ig{Comment the paragraph}
%CARLA offers a packaged version with additional assets that can be  imported into the package. Nonetheless, advanced customization and development options require the use of the build version of CARLA for both Windows and Linux.

The main features for the insertion of real agent behaviours in the simulation are based on five points (depicted in Fig. \ref{fig_Block_Diagram}): 1) \textbf{Avatar control}: from the CARLA's blueprint library that collects the architecture of all its actors and attributes, we modify the pedestrian blueprints to create an immersive and maneuverable VR interface between the person and the virtual world; 2) \textbf{Body tracking}: we use a set of inertial sensors and proprietary external software to capture the subject's motion through the real scene, and \textbf{motion perception}: we integrate the avatar's motion into the simulator via \emph{.bvh} files; 3) \textbf{Sound design}: since CARLA is an audio-less simulator, we introduce positional sound into the environment to enhance the immersion of the person; 4) \textbf{eHMI integration}: enabling the communication of state and intent information from autonomous vehicles to address interaction studies; 5) \textbf{Scenario simulation}: we design traffic scenarios within CARLA client, controlling the behaviour of vehicles and pedestrians.

%\ig{Subsection place has changed.}

\subsection{Avatar Control}

CARLA's blueprints (that include sensors, static actors, vehicles and walkers) have been specifically designed to be managed through the Python client API. Vehicles are actors that incorporate special internal components that simulate the physics of wheeled vehicles and can be driven by functions that provide driving commands (such as throttle, steering or braking). Walkers are operated in a similar way and its behavior can be directed by a controller. However, they are not designed to adopt real behaviors and therefore do not include wide range of unexpected movements and an immersive interface. 

We modify a walker blueprint to make an IK setup (inverse kinematics) for VR (full body room-scale) to support this. The tools available to capture the real human movement are the following: \textit{a) Oculus Quest 2} (for head tracking and user position control), and \textit{b) Motion controllers} (for both hands tracking). Oculus Quest 2 already has a safety distance system, delimiting the playing area through which the performer can move freely. The goal is to allow the performer to move within the safety zone that s/he has established, corresponding to a specific area of CARLA's scenario. 

The first step to achieve immersion is to modify the blueprint of the walker by attaching a virtual camera to its head to get a first-person feeling. The camera functions as the spectator's vision, while the skeletal mesh defines the appearance the walker adopts.
The displacement and perspective of the walker are activated, from certain minimum thresholds, with the translation and rotation of the VR headset, which are applied to the entire skeletal mesh. In addition, the VR glasses and both motion controllers adapt the pose of the neck and hand in real time. 
%To attend walker displacement, we wait for the performer to move the glasses. Once the distance moved or rotated by the performer exceeds certain thresholds, the same translation and rotation is applied to the entire skeletal mesh. In addition, Quest 2 and both motion controllers modify neck and hands pose respectively in real time. 

VR immersion for pedestrians is provided by implementing a head-mounted display (HMD) and creating an avatar in UE4. %The rotation and translation of the glasses are linked to the avatar movement as well as neck and hands tracking are performed. 
The subject wearing the VR glasses and controlling the avatar has complete freedom of movement within the preset area where we conduct the experiments.

\subsection{Full-Body Tracking}
%To integrate avatar control, just head and hands tracking is conducted to capture the real human movement within the simulator. However, full-body tracking is needed to collect enough information for autonomous vehicles to feed their trajectory prediction algorithms. 

To control the avatar (i.e., to represent the movement and pose of the subject inside the simulator), head (VR headset) and hand (motion controllers) tracking are not enough. Some kind of motion capture (MoCap) system is needed. There are multiple options, including vision-based systems with multiple cameras and inertial measurement units \cite{Menolotto2020}. In our case, an inertial wireless sensor system has been chosen, namely the Perception Neuron Studio (PNS) motion capture system \cite{PNS2022}, as a compromise solution between accuracy and usability. Each MoCap system includes a set of inertial sensors and straps that can be put on the joints easily, as well as a software for calibrating and capturing precise motion data. 
%To attend this, we use Perception Neuron motion capture systems. Perception Neuron offers a series of motion capture solutions that work with a series of inertial sensors and straps that can be put on the joints easily. Each mocap system includes software for calibrating and capturing precise motion data. Perception Neuron 
It also allows integration with other 3D rendering and animation software, such as iClone, Blender, Unity or UE4.

\subsection{Sound Design}
Other techniques for total immersion, such as sound design and real-world isolation, are other important aspects of behavioral research. World audio is absent in the CARLA simulator, but it is essential for interaction with the environment, as humans use spatial sound cues to track the location of other actors and predict their intentions.

We incorporate positional audio into the simulator, attaching a sound to each agent participating in the scene. Walking and talking sounds are assigned for pedestrians, and engine sounds parameterized by throttle and brake for vehicles. In addition, we add environmental sounds such as birds singing, wind and traffic sounds.

\subsection{External Human-Machine Interfaces (eHMI)}
\begin{figure}[t]
  \centering
  \includegraphics[width=\linewidth]{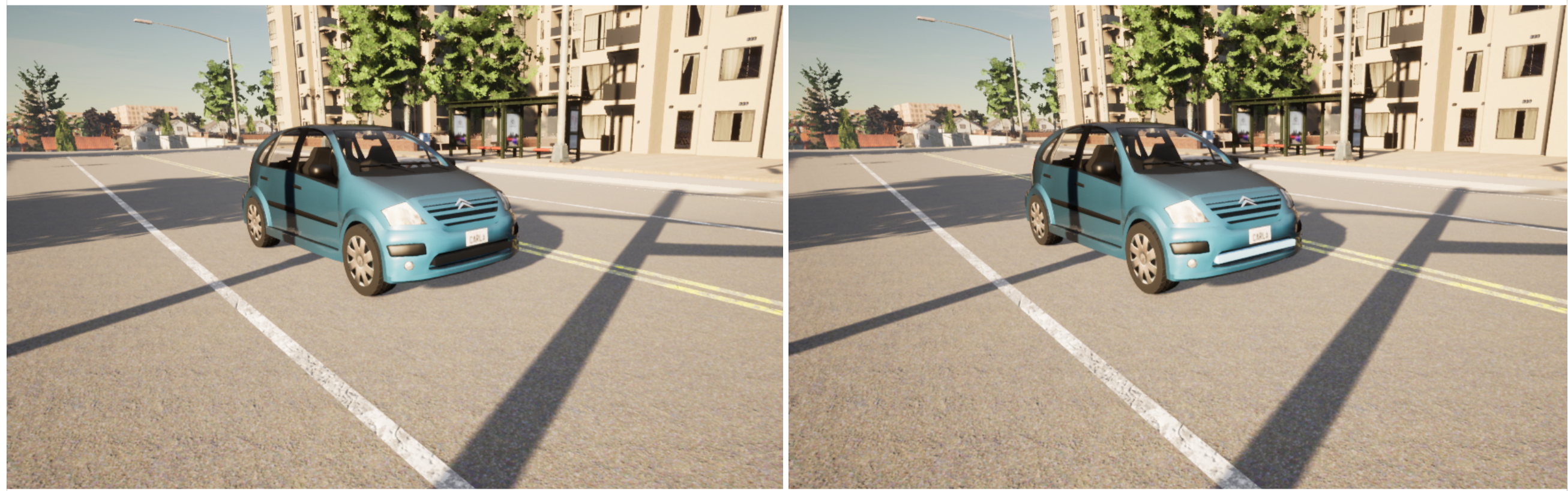}
  \caption{Left: vehicle without eHMI. Right: vehicle with eHMI activated.}
  \label{fig:eHMI}
\end{figure}

Communication between road users is an essential factor in traffic environments. In our experiments we include external human-machine interfaces (eHMI) for autonomous vehicles to communicate their status and intentions to the actual road user. The proposed eHMI design consists of a light strip along the entire front of the car, as depicted in Fig. \ref{fig:eHMI}. This allows studying the influence of the interface on decision making when the pedestrian's trajectory converges with the one followed by the vehicle in the virtual scenario.

\subsection{Traffic Scenario Simulation}
CARLA offers different options to simulate traffic and specific traffic scenarios. We use the Traffic Manager module to populate a simulation with realistic urban traffic conditions. The control of each vehicle is carried out in a specific thread. Communication with other layers is managed via synchronous messaging.

We can control traffic flow by setting parameters that enforce specific behaviors. For example, cars can be allowed to exceed speed limits, ignore traffic light conditions, ignore pedestrians, or force lane changes.

The subject is integrated into the simulator on a map that includes a 3D model of a city. Each map is based on an OpenDRIVE file that describes the fully annotated road layout. This feature allows us to design our own maps and reproduce the same traffic scenarios in real and virtual environments, to evaluate the integration of real behavior in the simulator and to be able to carry out presence studies by comparing the results of the interaction.

%The real road user with virtual vision and the rest of the actors navigate through a map that includes a 3D model of a town and its road definition. Every map is based on an OpenDRIVE file describing the road layout fully annotated. We will use this feature to design our own maps and reproduce the same traffic scenarios in real and virtual environments, to evaluate the integration of real behavior in the simulator by comparing interaction results.

\subsection{Recording, Playback and Motion Perception}
When running experiments, simulation data must be recorded and replayed for later analysis. CARLA has a native record and playback system that serializes the world information in each simulator tick for post-simulation recreation. However, this is only intended for tracking actors managed by the Python API and does not include the subject avatar or motion sensors. 

%When carrying out experiments, the data from simulation must be recorded and replayed for further analysis. CARLA features a native recorder and replayer system that serializes world information on every simulator tick for post simulation reenactment, but this is only intended to track actors managed by Python API (it does not include spectator avatar or Perception Neuron sensors). 

%Alongside the CARLA world state recording, recording and replaying real human full-body motion is essential for behaviour research. In this case, we use Axis Studio (Perception Neuron software) to record the whole body motion during the experiments when the performer has to interact with the traffic scene in virtual reality. Then, the recording is exported in a .bvh file which is integrated into the Unreal Engine editor.

Along with the recording of the state of the CARLA world, in our case the recording and playback of the actual full-body motion of the subject is essential. In our approach we use the Axis Studio software provided by Perception Neuron to record the body motion during experiments where the subject has to interact with the traffic scene in virtual reality. The recording is then exported in a \emph{.bvh} file which is subsequently integrated into the UE4 editor.

\begin{figure*}[ht]
  \centering
  \includegraphics[width=0.6\linewidth]{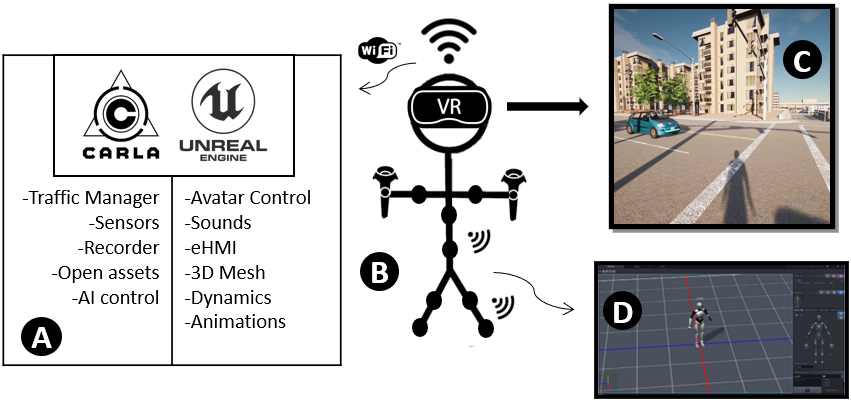}
  \caption{System Schematics. (A) Simulator CARLA-UE4. (B) VR headset, motion controllers and PN Studio sensors. (C) Spectator View in Virtual Reality. (D) Full-body tracking in Axis Studio.}
  \label{figure_2}
\end{figure*}

%Once interaction is registered in a recorder file, simulation is played back so that simulated sensors of the autonomous vehicles (i.e., radar, LiDAR, cameras) not only perceive the avatar skeletal mesh and its displacement, but also perceive the pose of all of its joints. 

Once the interaction is recorded, the simulation is played back so that the simulated sensors of the autonomous vehicles (i.e., radar, LiDAR, cameras) not only perceive the avatar's skeletal mesh and its displacement, but also the specific pose of all its joints (i.e., body language).

% REVISAR DESDE AQUÍ [10 MAY 2022]

\section{\uppercase{System implementation and Results}}

The overall scheme of the system is depicted in Fig. \ref{figure_2}. In the following we describe the hardware and software configurations and present preliminary results. 

\subsection{Hardware setup}
\begin{figure}[t]
  \centering
  \includegraphics[width=\linewidth]{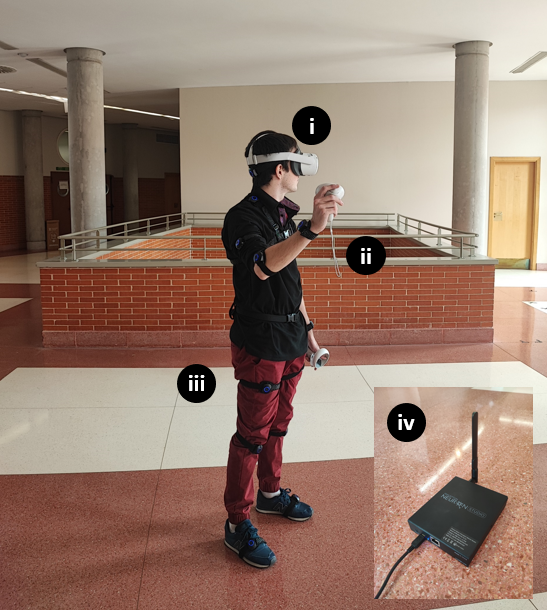}
  \caption{Hardware setup. (i) VR headset (Quest 2): transfer the image from the environment to the performer. (ii) Motion controllers: allow control of the avatar's hands. (iii) PN Studio sensors: provide body tracking withstanding magnetic interference. (iv) Studio Transceiver: receives sensors data wirelessly by 2.4GHz.}
  \label{figure_3}
\end{figure}

The overall hardware setup is shown in Fig. \ref{figure_3}. During experiments, we use the Oculus Quest 2 as our head-mounted device (HMD), created by Meta, which has 6GB RAM processor, two adjustable 1832 x 1920 lenses, 90Hz refresh rate and an internal memory of 256 GB. Quest 2 features WiFi 6, Bluetooth 5.1, and USB Type-C connectivity, SteamVR support and 3D speakers. For full-body tracking we use PNS packaged solution with inertial trackers. The kit includes standalone VR headset, 2 motion controllers, 17 Studio Inertial Body sensors, 14 set of straps, 1 charging case and 1 Studio Transceiver.

To conduct the experiments, we reserved a preset area wide enough and free of obstacles where the subject can act as a real pedestrian inside the simulator. Quest 2 and motion controllers are connected to PC via Oculus link or WiFi as follows:

\begin{itemize}
\item Wired connection: via the Oculus Link cable or other similar high quality USB 3. 
\item Wireless connection: via WiFi by enabling Air Link from the Meta application, or using Virtual Desktop and SteamVR.
\end{itemize}

The subject puts on the straps of the appropriate length and places the PN Studio sensors into the bases. 
The transceiver is attached to the PC via USB. After launching the build version of CARLA for Windows, Quest 2 enables "VR Preview" in the UE4 editor.

\subsection{Software setup}
VR Immersion System is currently dependent on UE4.24 and Windows 10 OS due to CARLA build, and Quest 2 Windows-only dependencies. Using TCP socket plugin, all the actor locations and other useful parameters for the editor are sent from the Python API to integrate, for example, the sound of each actor or the eHMI of the autonomous vehicle. "VR Preview" launches the game on the HMD. 

Perception Neuron Studio works with Axis Studio which supports up to 3 subjects at a time while managing up to 23 sensors for the body and fingers.

\begin{figure}[t]
  \centering
  \includegraphics[width=\linewidth]{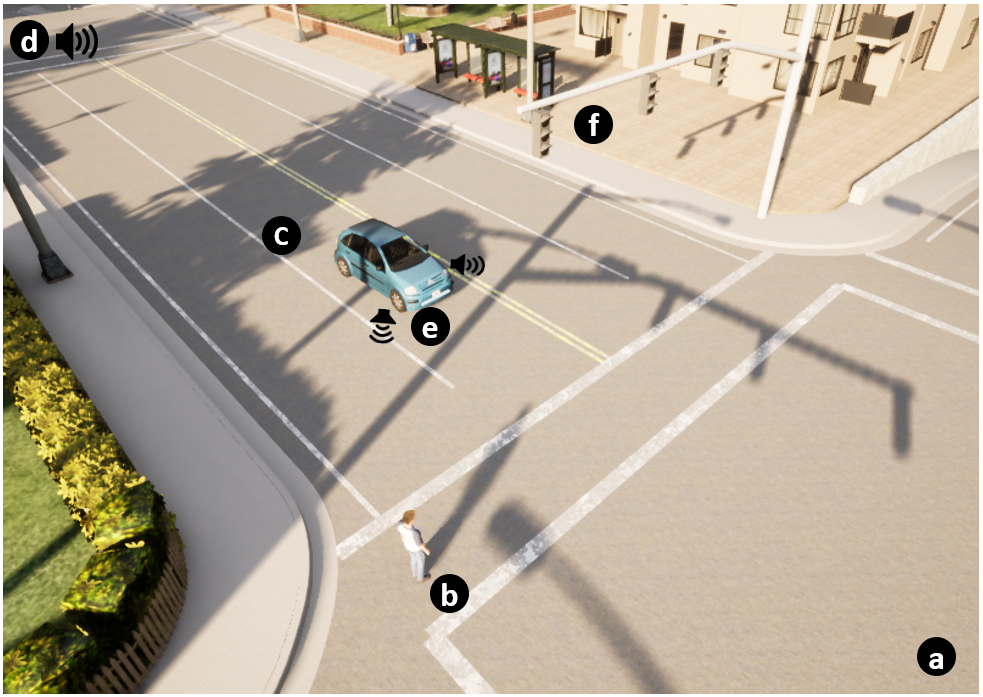}
  \caption{Simulation of Interactive traffic situations. (a) 3D world design. (b) Pedestrian matches the performer avatar. (c) Autonomous vehicle. (d) Environment sounds and agents sounds. (e) eHMI. (f) Traffic lights and traffic signs.}
  \label{figure_4}
\end{figure}

\subsection{Results}
The safety of automated driving functions depends highly on reliable corner cases detection. A recent research work describes corner cases (CC) as data that occur infrequently, represent a dangerous situation and are sparse in the datasets \cite{cornercases}. We aim to integrate behaviours of real subjects in the generation of wide variety of scenarios with pedestrians involved, including CC, in order to provide training data with real-time interaction. The preliminary scenario designed includes the interaction between a single autonomous vehicle and a single pedestrian. The vehicle circulates on the road when it reaches a pedestrian crossing where the pedestrian shows clear intentions to cross. The pedestrian receives information on status and vehicle intentions through an eHMI and can track its location by listening to the engine sound. Lighting and weather conditions are favorable (sunny day).

\begin{figure}[t]
  \centering
  \includegraphics[width=\linewidth]{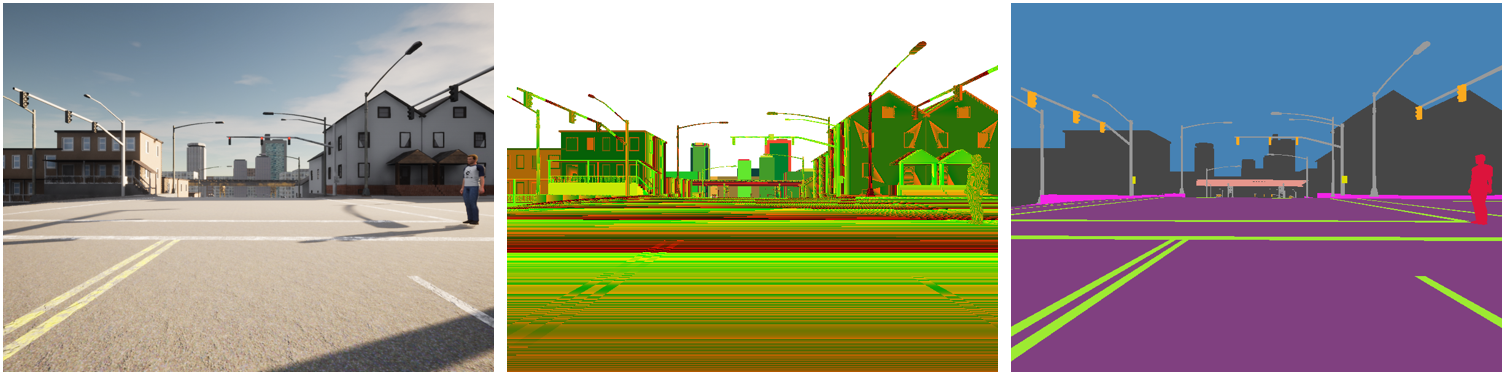}
  \caption{Cameras RGB / Depth / Segmentation.}
  \label{figure_5}
\end{figure}

\begin{figure}[t]
  \centering
  \includegraphics[width=\linewidth]{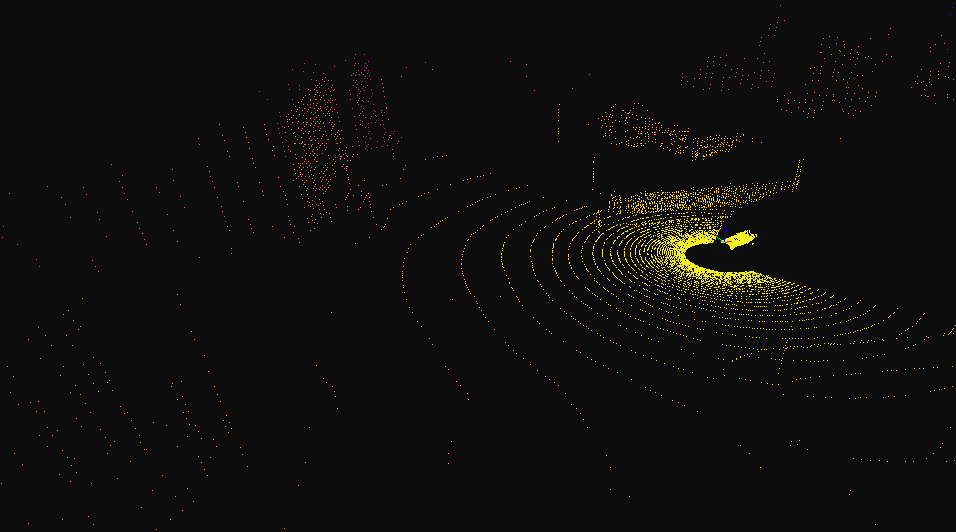}
  \caption{LiDAR point cloud (ray-casting).}
  \label{figure_6}
\end{figure}

It is possible to spot the vehicle a few metres ahead of the pedestrian crossing before starting the action. When the pedestrian decides whether or not to wait for the vehicle to come to a complete stop, s/he pays attention to the speed and deceleration of the vehicle before it reaches the intersection. Sensors attached to the vehicle capture the image of the scene as shown in Fig. \ref{figure_5}, and detect the pedestrian as shown in LiDAR point cloud in Fig. \ref{figure_6}.

As a preliminary usability assessment, six real subjects completed a 15-item presence scale consisting in five items for self-presence, five items for autonomous vehicle presence and five items for environmental presence. The questions raised are depicted in Appendix A. In addition, we request subjective feedback about system usability and user experience.

Most of the participants felt a strong self-presence (M=3.97, SD=0.528), fitting into the avatar's body. Nevertheless, some of them were skeptical of the damage the avatar was exposed to (such as being run over) and did not feel under any real threat. Regarding autonomous vehicle presence  (M=3.60, SD=0.657), the engine's sound composed the point of greatest disagreement among the participants, since some found it very useful to locate the vehicle, while others did not even notice or found it annoying. The image of the vehicle did not look completely clear-cut until the vehicle was close enough to the avatar, and its stop in front of the pedestrian cross did not feel threatening in the sense that its braking was appreciated too conservative. The environmental presence (M=4.63, SD=0.266) was felt very strongly by all the participants that felt really involved in an urban environment and commented that adding other agents or more traffic on the road would enhance the experience.

VR Immersion allows to integrate real participants in interactive situations on the road and to use objective metrics (such as collisions and reaction times) in order to measure interaction quality without real danger. CARLA enables custom traffic scenarios while UE4 provides the immersive and interactive interface into the same simulator.

\section{\uppercase{Conclusions}}
In this paper, we have presented a preliminary framework to enable real-time interaction between real agents and the simulated environment. At this level, more results are expected from testing interactive traffic scenes with multiple real participants. 

Regarding the immersive sensation, the virtual environment displayed from the glasses varies between 15 and 20 fps, showing a stable image to the performer that allows him/her to interact with the world of CARLA. The wireless connection is prioritized over the wired connection, since it allows a greater degree of freedom of movement. The performer pose within the virtual environment is registered by Perception Neuron, generating useful sequences to train and validate predictive models to, for example, predict future actions and trajectories of traffic agents.

The scenario shown in this paper serves as a basis for a preliminary presentation of the proposed methodology. From here, as future work, it is intended, on the one hand, to carry out a study of presence, interaction and communication with equivalent real and virtual environments. And on the other hand, to develop a data recording campaign with multiple subjects, different traffic scenarios and varied environmental conditions, in order to generate a dataset that can be used to train action and trajectory prediction models. 

\section*{\uppercase{Acknowledgements}}
This work was mainly supported by the Spanish Ministry of Science and Innovation (Research Grant PID2020-114924RB-I00) and in part by the Community Region of Madrid (Research Grant 2018/EMT-4362 SEGVAUTO 4.0-CM).

\bibliographystyle{apalike}
{\small
\bibliography{chira22}}

\begin{thebibliography}{}

\bibitem[Bhagavathula et~al., 2018]{Bhagavathula2018}
Bhagavathula, R., Williams, B., Owens, J., and Gibbons, R. (2018).
\newblock The reality of virtual reality: A comparison of pedestrian behavior
  in real and virtual environments.
\newblock {\em Proceedings of the Human Factors and Ergonomics Society Annual
  Meeting}, 62(1):2056--2060.

\bibitem[Bogdoll et~al., 2021]{cornercases}
Bogdoll, D., Breitenstein, J., Heidecker, F., Bieshaar, M., Sick, B.,
  Fingscheidt, T., and Z\"{o}llner, J.~M. (2021).
\newblock Description of corner cases in automated driving: Goals and
  challenges.
\newblock arXiv.

\bibitem[Bruder et~al., 2009]{Bruder2009}
Bruder, G., Steinicke, F., Rothaus, K., and Hinrichs, K. (2009).
\newblock Enhancing presence in head-mounted display environments by visual
  body feedback using head-mounted cameras.
\newblock In {\em 2009 International Conference on CyberWorlds}, pages 43--50.

\bibitem[Chang et~al., 2017]{Chang2017}
Chang, C.-M., Toda, K., Sakamoto, D., and Igarashi, T. (2017).
\newblock Eyes on a car: an interface design for communication between an
  autonomous car and a pedestrian.
\newblock In {\em Proceedings of the 9th International Conference on Automotive
  User Interfaces and Interactive Vehicular Applications (AutomotiveUI'17)},
  pages 65--73.

\bibitem[Deb et~al., 2020]{Shuchisnigdha2020}
Deb, S., Carruth, D.~W., Fuad, M., Stanley, L.~M., and Frey, D. (2020).
\newblock Comparison of child and adult pedestrian perspectives of external
  features on autonomous vehicles using virtual reality experiment.
\newblock In Stanton, N., editor, {\em Advances in Human Factors of
  Transportation}, pages 145--156. Springer International Publishing.

\bibitem[Deb et~al., 2017]{Deb2017}
Deb, S., Carruth, D.~W., Sween, R., Strawderman, L., and Garrison, T.~M.
  (2017).
\newblock Efficacy of virtual reality in pedestrian safety research.
\newblock {\em Applied Ergonomics}, 65:449--460.

\bibitem[{DeepDrive}, 2018]{deepdrive}
{DeepDrive} (2018).
\newblock Deepdrive.
\newblock https://deepdrive.io/ (accessed on 22 April 2022).

\bibitem[Doric et~al., 2016]{Doric2016}
Doric, I., Frison, A.-K., Wintersberger, P., Riener, A., Wittmann, S.,
  Zimmermann, M., and Brandmeier, T. (2016).
\newblock A novel approach for researching crossing behavior and risk
  acceptance: The pedestrian simulator.
\newblock In {\em Proceedings of the 8th International Conference on Automotive
  User Interfaces and Interactive Vehicular Applications (AutomotiveUI'16)},
  pages 39--44.

\bibitem[Dosovitskiy et~al., 2017]{carla2017}
Dosovitskiy, A., Ros, G., Codevilla, F., Lopez, A., and Koltun, V. (2017).
\newblock {CARLA}: {An} open urban driving simulator.
\newblock In {\em Proceedings of the 1st Annual Conference on Robot Learning},
  pages 1--16.

\bibitem[Eady, 2019]{Eady2019}
Eady, T. (2019).
\newblock Simulations can't solve autonomous driving because they lack
  important knowledge about the real world - large-scale real world data is the
  only way.
\newblock Medium Self-Driving Cars.
\newblock
  https://medium.com/@trenteady/simulation-cant-solve-autonomous-driving-because-it-lacks-necessary-empirical-knowledge-403feeec15e0
  (accessed on 8 April 2022).

\bibitem[Farooq et~al., 2018]{Farooq2018}
Farooq, B., Cherchi, E., and Sobhani, A. (2018).
\newblock Virtual immersive reality for stated preference travel behavior
  experiments: A case study of autonomous vehicles on urban roads.
\newblock {\em Transportation Research Record}, 2672(50):35--45.

\bibitem[Fernández-Llorca and Gómez, 2021]{Llorca2021}
Fernández-Llorca, D. and Gómez, E. (2021).
\newblock Trustworthy autonomous vehicles.
\newblock {\em Publications Office of the European Union, Luxembourg,}, EUR
  30942 EN, JRC127051.

\bibitem[Gruenefeld et~al., 2019]{Gruenefeld2019}
Gruenefeld, U., Wei\ss{}, S., L\"{o}cken, A., Virgilio, I., Kun, A.~L., and
  Boll, S. (2019).
\newblock Vroad: Gesture-based interaction between pedestrians and automated
  vehicles in virtual reality.
\newblock In {\em Proceedings of the 11th International Conference on
  Automotive User Interfaces and Interactive Vehicular Applications: Adjunct
  Proceedings}, AutomotiveUI '19, page 399–404.

\bibitem[Holl\"{a}nder et~al., 2019]{Hollander2019}
Holl\"{a}nder, K., Wintersberger, P., and Butz, A. (2019).
\newblock Overtrust in external cues of automated vehicles: An experimental
  investigation.
\newblock In {\em Proceedings of the 11th International Conference on
  Automotive User Interfaces and Interactive Vehicular Applications}, page
  211–221, New York, NY, USA. Association for Computing Machinery.

\bibitem[Iryo-Asano et~al., 2018]{IryoAsano2018}
Iryo-Asano, M., Hasegawa, Y., and Dias, C. (2018).
\newblock Applicability of virtual reality systems for evaluating
  pedestrians’ perception and behavior.
\newblock {\em Transportation Research Procedia}, 34:67--74.

\bibitem[Kalra and Paddock, 2016]{Kalra2016}
Kalra, N. and Paddock, S.~M. (2016).
\newblock Driving to safety: how many miles of driving would it take to
  demonstrate autonomous vehicle reliability?
\newblock Research report, RAND Corporation.

\bibitem[L\"{o}cken et~al., 2019]{Locken2019}
L\"{o}cken, A., Golling, C., and Riener, A. (2019).
\newblock How should automated vehicles interact with pedestrians? a
  comparative analysis of interaction concepts in virtual reality.
\newblock In {\em Proceedings of the 11th International Conference on
  Automotive User Interfaces and Interactive Vehicular Applications},
  AutomotiveUI '19, page 262–274.

\bibitem[Martinez et~al., 2021]{Hernandez2021}
Martinez, A.~H., Díaz, J.~L., Daza, I.~G., and Llorca, D.~F. (2021).
\newblock Data-driven vehicle speed detection from synthetic driving simulator
  images.
\newblock In {\em 2021 IEEE International Intelligent Transportation Systems
  Conference (ITSC)}, pages 2617--2622.

\bibitem[Menolotto et~al., 2020]{Menolotto2020}
Menolotto, M., Komaris, D.-S., Tedesco, S., O’Flynn, B., and Walsh, M.
  (2020).
\newblock Motion capture technology in industrial applications: A systematic
  review.
\newblock {\em Sensors}, 20(19).

\bibitem[Nazemi et~al., 2021]{Nazemi2021}
Nazemi, M., {van Eggermond}, M., Erath, A., Schaffner, D., Joos, M., and
  Axhausen, K.~W. (2021).
\newblock Studying bicyclists’ perceived level of safety using a bicycle
  simulator combined with immersive virtual reality.
\newblock {\em Accident Analysis \& Prevention}, 151:105943.

\bibitem[Nguyen et~al., 2019]{Nguyen2019}
Nguyen, T.~T., Holl\"{a}nder, K., Hoggenmueller, M., Parker, C., and Tomitsch,
  M. (2019).
\newblock Designing for projection-based communication between autonomous
  vehicles and pedestrians.
\newblock In {\em Proceedings of the 11th International Conference on
  Automotive User Interfaces and Interactive Vehicular Applications},
  AutomotiveUI '19, page 284–294. Association for Computing Machinery.

\bibitem[Noitom, 2022]{PNS2022}
Noitom (2022).
\newblock Perception neuron studio syste.
\newblock https://neuronmocap.com/perception-neuron-studio-system (accessed on
  5 May 2022).

\bibitem[Nuñez~Velasco et~al., 2021]{NunezVelasco2021}
Nuñez~Velasco, J.~P., de~Vries, A., Farah, H., van Arem, B., and Hagenzieker,
  M.~P. (2021).
\newblock Cyclists' crossing intentions when interacting with automated
  vehicles: A virtual reality study.
\newblock {\em Information}, 12(1).

\bibitem[{Nuñez Velasco} et~al., 2019]{NunezVelasco2019}
{Nuñez Velasco}, J.~P., Farah, H., {van Arem}, B., and Hagenzieker, M.~P.
  (2019).
\newblock Studying pedestrians’ crossing behavior when interacting with
  automated vehicles using virtual reality.
\newblock {\em Transportation Research Part F: Traffic Psychology and
  Behaviour}, 66:1--14.

\bibitem[Ros et~al., 2016]{Ros2016}
Ros, G., Sellart, L., Materzynska, J., Vazquez, D., and Lopez, A.~M. (2016).
\newblock The synthia dataset: A large collection of synthetic images for
  semantic segmentation of urban scenes.
\newblock In {\em 2016 IEEE Conference on Computer Vision and Pattern
  Recognition (CVPR)}, pages 3234--3243.

\bibitem[Shah et~al., 2018]{airsim}
Shah, S., Dey, D., Lovett, C., and Kapoor, A. (2018).
\newblock Airsim: High-fidelity visual and physical simulation for autonomous
  vehicles.
\newblock In Hutter, M. and Siegwart, R., editors, {\em Field and Service
  Robotics}, pages 621--635. Springer International Publishing.

\bibitem[Silvera et~al., 2022]{Silvera2022}
Silvera, G., Biswas, A., and Admoni, H. (2022).
\newblock Dreyevr: Democratizing virtual reality driving simulation for
  behavioural \& interaction research.

\bibitem[Ullmann et~al., 2020]{Ullmann2020}
Ullmann, D., Kreimeier, J., G\"{o}tzelmann, T., and Kipke, H. (2020).
\newblock Bikevr: A virtual reality bicycle simulator towards sustainable urban
  space and traffic planning.
\newblock In {\em Proceedings of the Conference on Mensch Und Computer}, MuC
  '20, page 511–514. Association for Computing Machinery.

\bibitem[Vázquez et~al., 2014]{Vazquez2014}
Vázquez, D., López, A.~M., Marín, J., Ponsa, D., and Gerónimo, D. (2014).
\newblock Virtual and real world adaptation for pedestrian detection.
\newblock {\em IEEE Transactions on Pattern Analysis and Machine Intelligence},
  36(4):797--809.

\bibitem[Weng et~al., 2021]{Weng2020_AIODrive}
Weng, X., Man, Y., Park, J., Yuan, Y., Cheng, D., O'Toole, M., and Kitani, K.
  (2021).
\newblock {All-In-One Drive: A Large-Scale Comprehensive Perception Dataset
  with High-Density Long-Range Point Clouds}.
\newblock {\em arXiv}.

\bibitem[Xu et~al., 2017]{Xu2017}
Xu, J., Lin, Y., and Schmidt, D. (2017).
\newblock Exploring the influence of simulated road environments on cyclist
  behavior.
\newblock {\em The International Journal of Virtual Reality}, 17(3):15--26.

\end{thebibliography}

\section*{\uppercase{Appendix A}}

\subsection*{Self-presence Scale items}

To what extent did you feel that… (1= not at all – 5 very strongly)

\begin{enumerate}
\item You could move the avatar's hands.
\item The avatar's displacement was your own displacement.
\item The avatar's body was your own body.
\item If something happened to the avatar, it was happening to you.
\item The avatar was you.
\end{enumerate}

\subsection*{Autonomous vehicle presence Scale items} 

To what extent did you feel that… (1= not at all – 5 very strongly)

\begin{enumerate}
\item The vehicle was present.
\item The vehicle dynamics and its movement were natural.
\item The sound of the vehicle helped you to locate it.
\item The vehicle was aware of your presence.
\item The vehicle was real.
\end{enumerate}

\subsection*{Environmental presence Scale items} 

To what extent did you feel that… (1= not at all – 5 very strongly)

\begin{enumerate}
\item You were really in front of a pedestrian crossing.
\item The road signs and traffic lights were real.
\item You really crossed the pedestrian crossing.
\item The urban environment seemed like the real world.
\item It could reach out and touch the objects in the urban environment.
\end{enumerate}

%If any, the appendix should appear directly after the
%references without numbering, and not on a new %page. To do so please use the following command:
%\textit{$\backslash$section*\{APPENDIX\}}

\end{document}